\documentclass{article}

\newenvironment{tightlist}{
\begin{list}{$\bullet$}{
    \setlength{\topsep}{.1em}
    \setlength{\partopsep}{0in}
    \setlength{\parskip}{0in}
    \setlength{\itemsep}{0in}
    \setlength{\parsep}{0in}
    \setlength{\leftmargin}{1em}
    \setlength{\rightmargin}{0in}
    \setlength{\itemindent}{0in}
}}
{\end{list}}

\usepackage{adjustbox}

\usepackage[table]{xcolor}

\usepackage{wrapfig}
\usepackage{tikz}
\usetikzlibrary{calc, shapes, arrows, positioning}
\usetikzlibrary{bayesnet}
\usepackage{microtype}
\usepackage{afterpage} 

\usepackage{tcolorbox} 
\usepackage{amsmath} 
\usepackage{graphicx}
\usepackage{booktabs} 

\usepackage{hyperref}

\usepackage{amsmath}

\usepackage{subcaption}
\newcommand{\todo}[1]{\textcolor{red}{}}

\usepackage{amsthm}
\usepackage{algorithm}
\usepackage{algpseudocode}
\usepackage{amsfonts}
\newcommand{\Figref}[1]{Figure~\ref{#1}}
\newcommand{\figref}[1]{Figure~\ref{#1}}

\newcommand{\secref}[1]{Section~\ref{#1}}
\newcommand{\Algref}[1]{Algorithm~\ref{#1}}
\renewcommand{\algref}[1]{Algorithm~\ref{#1}}
\newcommand{\appref}[1]{Appendix~\ref{#1}}

\usepackage[preprint]{neurips_2025}

\usepackage[utf8]{inputenc} 
\usepackage[T1]{fontenc}    
\usepackage{hyperref}       
\usepackage{url}            
\usepackage{booktabs}       
\usepackage{amsfonts}       
\usepackage{nicefrac}       
\usepackage{microtype}      

\newtheorem{theorem}{Theorem}

\author{%
  Isha Puri$^{1}$ \\
  \And
  Shiv Sudalairaj$^{2}$ \\
  \And
  GX Xu$^{2}$ \\
  \And
  Kai Xu$^{2}$ \\
  \And
  Akash Srivastava$^{2}$\\
  \And 
   \textsuperscript{1}MIT CSAIL \hspace{.5cm} \textsuperscript{2}RedHat AI Innovation
}
\begin{document}

\title{Rollout Roulette: A Probabilistic Inference Approach to Inference-Time Scaling of LLMs using Particle-Based Monte Carlo Methods}

\maketitle

\begin{abstract}
Large language models (LLMs) have achieved significant performance gains via scaling up model sizes and/or data. 
However, recent evidence suggests diminishing returns from such approaches, motivating a pivot to scaling test-time compute.
Existing deterministic inference-time scaling methods, usually with reward models, cast the task as a search problem, but suffer from a key limitation: early pruning. Due to inherently imperfect reward models, promising trajectories may be discarded prematurely, leading to suboptimal performance. We propose a novel inference-time scaling approach by adapting particle-based Monte Carlo methods. Our method maintains a diverse set of candidates and robustly balances exploration and exploitation. 
Our empirical evaluation demonstrates that our particle filtering methods have a 4--16x better scaling rate over deterministic search counterparts on both various challenging mathematical and more general reasoning tasks. 
Using our approach, we show that Qwen2.5-Math-1.5B-Instruct surpasses GPT-4o accuracy in only 4 rollouts, while Qwen2.5-Math-7B-Instruct scales to o1 level accuracy in only 32 rollouts.
Our work not only presents an effective method to inference-time scaling, but also connects rich literature in probabilistic inference with inference-time scaling of LLMs to develop more robust algorithms in future work. Code and further information is available at \url{https://probabilistic-inference-scaling.github.io/} 

\end{abstract}

\section{Introduction}

Large language models (LLMs) continue to improve through larger architectures and massive training corpora \citep{kaplan2020scalinglawsneurallanguage,snell2024scalingllm}.
In parallel, inference-time scaling (ITS)---allocating more computation at inference time---has emerged as a complementary approach to improve performance without increasing model size. 
Recent work has shown that ITS enables smaller models to match or exceed the accuracy of larger ones on complex reasoning tasks \citep{beeching2024scalingtesttimecompute}, with proprietary systems like OpenAI's o1 and o3 explicitly incorporating multi-trial inference for better answers \citep{openai2024openaio1card}.

\begin{figure}[h]
    \centering
    \includegraphics[width=\linewidth]{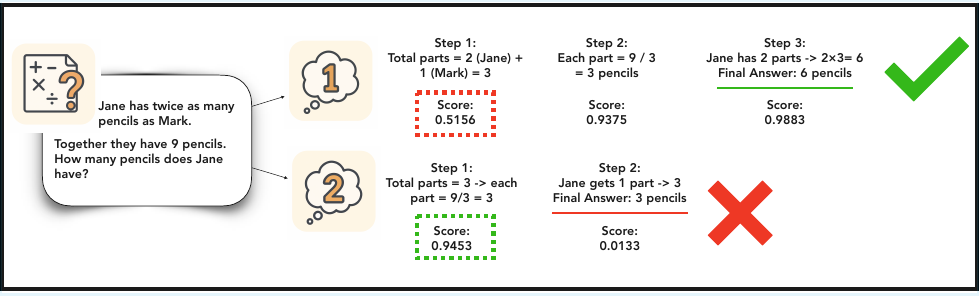}
    \caption{A true example of PRM assigning a lower score to the first step of a solution that turns out to be correct. In deterministic scaling methods, this solution would have been discarded in favor for one that had a higher initial PRM score but turned out to be incorrect.}
    \label{fig:answerFallingOffBeam}
    \vspace{-1em}
\end{figure}

Beyond answer-level scaling methods like self-consistency and best-of-$n$ sampling \citep{brown2024largelanguage},
a popular class of ITS methods formulates inference as search guided by a process reward model (PRM), which scores partial sequences step-by-step. 
This perspective has led to the adoption of algorithms like beam search \citep{zhou2024languageagenttreesearch} and Monte Carlo tree search \citep{guan2025rstarmathsmall}.
These methods refine LLM outputs by prioritizing trajectories that score highly according to the PRM. However, they also inherit a major limitation from classical search: early pruning. Because the PRM is an imperfect approximation of true correctness, these methods often eliminate promising candidates too early-based on noisy or miscalibrated partial scores. This failure mode is illustrated in \figref{fig:answerFallingOffBeam}. The PRM assigns a higher initial score to Answer 2, which ultimately turns out to be incorrect, and a lower score to the correct Answer 1. A deterministic method like beam search would discard Answer 1 after the first step, never recovering it—even though it is the correct solution. Such brittleness is inherent to greedy search: once a path is pruned, it cannot be revisited.

To address this limitation, we instead maintain a distribution over the possible paths during generation to represent the uncertainty we would like to account for due to the imperfection in reward modeling and propagate it through sampling.
We realize this approach in a probabilistic inference framework for inference-time scaling, in which we use particle filtering to sample from the posterior distribution over accepted trajectories. 
Our method maintains a weighted population of candidate sequences that evolves over time, balancing exploitation of high-PRM paths with stochastic exploration of alternatives. Unlike beam search, which targets the mode of an approximate distribution, particle filtering samples from the typical set, making it inherently more robust to noise and multi-modality in the reward landscape. 
High-scoring candidates are favored but not allowed to dominate, ensuring that low-probability (but potentially correct) paths remain in play.

We demonstrate that this simple shift-from deterministic search to sampling with uncertainty---produces strong empirical gains. On mathematical and reasoning tasks, our method significantly outperforms existing ITS approaches across multiple model families. For instance, using Qwen2.5-Math-1.5B-Instruct and a compute budget of 4, our method surpasses GPT-4o; with a budget of 32, the 7B model surpasses o1 accuracy.

Our key contributions are as follows.
\begin{enumerate}
    \item We propose a particle filtering algorithm for inference-time scaling that mitigates early pruning by maintaining exploration across trajectories with uncertainty. We formulate ITS as posterior inference over a state space model defined by an LLM (transition model) and PRM (emission model), enabling principled application of probabilistic inference tools.
    \item We present strong results on mathematical and out-of-domain reasoning tasks and study Particle Filtering's scaling performance.
    \item We ablate compute allocation strategies (parallel vs. iterative), PRM aggregation methods, and generation temperature, proposing a new model-based reward aggregation method that improves stability and performance.
    \item We demonstrate that our proposed methods have 4--16x faster scaling speed than previous methods based on a search formulation on the MATH500 and AIME 2024 datasets, with small language models in the Llama and Qwen families. We show that PF can scale Qwen2.5-Math-1.5B-Instruct to surpasses GPT-4o accuracy with only a budget of 4 and scale Qwen2.5-Math-7B-Instruct to o1 accuracy with a budget of 32.
\end{enumerate}

\section{Background}\label{sec:bg}


\textbf{State space models} describe sequential systems that evolve stepwise, typically over time \citep{Särkkä_2013}. 
They consist of a sequence of latent states $\{x_t\}_{t=1}^T$ and corresponding observations $\{o_t\}_{t=1}^T$
The evolution of states is governed by a transition model $p(x_t \mid x_{<t-1})$, and the observations are governed by the emission model $p(o_t | x_t)$.
The joint distribution of states and observations is given by:
$
    p(x_{1:T}, o_{1:T}) = p(x_1) {\prod}_{t=2}^T p(x_t \mid x_{<t-1}) {\prod}_{t=1}^T p(o_t \mid x_t)
$,
where $p(x_1)$ is the prior.

\textbf{Probabilistic inference} in SSMs involves estimating the posterior distribution of the hidden states given the observations, $p(x_{1:T} | o_{1:T})$ \citep{Särkkä_2013}, which is generally intractable due to the high dimensionality of the state space and the dependencies in the model. 
Common approaches approximate the posterior through sampling-based methods or variational approaches \citep{mackay2003information}. \textbf{Particle filtering} (PF) is a sequential Monte Carlo method to approximate the posterior distribution in SSMs \citep{nonlinearfiltering,sequentialmonte}. 
PF represents the posterior using a set of $N$ weighted particles $\{x_t^{(i)}, w_t^{(i)}\}_{i=1}^N$, where $x_t^{(i)}$ denotes the $i^\text{th}$ particle at time $t$, and $w_t^{(i)}$ is its associated weight. 
The algorithm iteratively propagates particles using the transition model 
and updates weights based on the emission model:
$
    w_t^{(i)} \propto w_{t-1}^{(i)} p(o_t \mid x_t^{(i)})
$.

\section{Method}\vspace{-1.5em}

\begin{figure}[h]
    \centering
    \begin{adjustbox}{center}
        \includegraphics[width=1.0\linewidth]{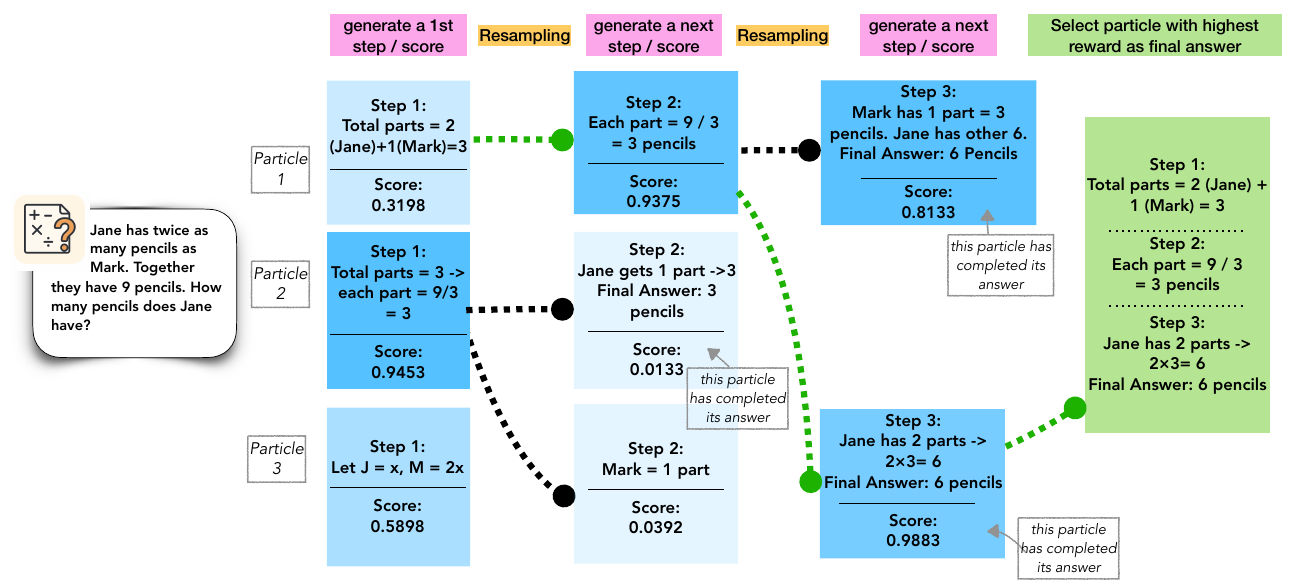}
    \end{adjustbox}
    \caption{Inference-time scaling with particle filtering: initialize $n$ particles, generate a step for each, score with the PRM, resample via softmax-weighted scores, and repeat until full solutions are formed.}
    \label{fig:main-method-figure}
\end{figure}

We formulate inference-time scaling for a language model $p_M$ as approximate posterior inference in a state space model (SSM) defined over token sequences. Let $x_{1:T}$ denote the sequence of generated tokens (or chunks, such as steps in a math problem), and let $o_{1:T}$ denote binary observations indicating whether the steps so far are accepted, given a prompt $c$.

The forward model defines the joint distribution over latent states and observations as:

\begin{equation}\label{eq:target-gt}
\begin{aligned}
p_M(x_{1:T}, o_{1:T} \mid c) \propto \prod_{t=1}^T p_M(x_t \mid c, x_{<t-1}) \prod_{t=1}^T p(o_t \mid c, x_{<t}),
\end{aligned}
\end{equation}

where the transition model $p_M(x_t \mid c, x_{<t-1})$ is given by the language model $M$, and the emission model $p(o_t \mid c, x_{<t})$ is a Bernoulli distribution: 
$p(o_t \mid c, x_{<t}) = \mathcal{B}(o_t ; r(c, x_t))$,
with reward function $r(c, x_t)$ encoding the acceptability of the step $x_t$. \Figref{fig:plate} illustrates this SSM.

Given this model, our goal is to infer the posterior distribution over latent trajectories that yield fully accepted sequences, i.e., $p_M(x_{1:T} \mid c, o_{1:T} = \mathbf{1})$. This formulation makes particle filtering a natural and theoretically grounded choice for inference.

In practice, the true reward function $r$ is often unavailable. Following prior work, we approximate it using a pre-trained preference or reward model (PRM) $\hat{r}$ suited to the target domain (e.g., math reasoning). This results in an approximate emission model:
$\hat{p}(o_t \mid c, x_{<t}) = \mathcal{B}(o_t ; \hat{r}(c, x_{<t}))$.
Substituting this into the forward model, we obtain the approximate joint distribution:

\begin{equation}\label{eq:target}
\begin{aligned}
\hat{p}_M(x{1:T}, o_{1:T} \mid c) \propto \prod_{t=1}^T p_M(x_t \mid c, x_{<t-1}) \prod_{t=1}^T \hat{p}(o_t \mid c, x_{<t}),
\end{aligned}
\end{equation}

and correspondingly aim to estimate the posterior $\hat{p}_M(x{1:T} \mid c, o_{1:T} = \mathbf{1})$.

\begin{figure}[t]
    \centering
    \vspace{0em}
    \begin{tikzpicture}
        \node[latent] (x1) {$x_1$}; 
        \node[latent, right=of x1] (x2) {$x_2$}; 
        \node[right=0.3cm of x2] (dots) {$\cdots$}; 
        \node[latent, right=0.3cm of dots] (xT) {$x_T$}; 

        \node[const, above=of x1] (c) {$c$};

        \node[obs, below=of x1] (o1) {$o_1$}; 
        \node[obs, below=of x2] (o2) {$o_2$}; 
        \node[below=1.25cm of dots] (odots) {$\cdots$}; 
        \node[obs, below=of xT] (oT) {$o_T$}; 

        \node[below right=-0.1cm of x1] (x1p) {\tiny \color{blue} $p_M(x_1 \mid c)$};
        \node[below right=-0.1cm of x2] (x2p) {\tiny \color{blue} $p_M(x_2 \mid c, x_1)$};
        \node[below right=-0.1cm of xT] (xTp) {\tiny \color{blue} $p_M(x_T \mid c, x_{<T})$};
        \node[below right=-0.1cm of o1] (o1p) {\tiny \color{red} $\hat{p}(o_1 \mid c, x_1)$};
        \node[below right=-0.1cm of o2] (o2p) {\tiny \color{red} $\hat{p}(o_2 \mid c, x_2)$};
        \node[below right=-0.1cm of oT] (oTp) {\tiny \color{red} $\hat{p}(o_t \mid c, x_{<t})$};
        \node[align=right, left=0.5cm of x1] (llm) {\small \color{blue} language\\\small \color{blue} model};
        \node[align=right, left=0.5cm of o1]  (rm) {\small \color{red} reward\\\small \color{red} model};

        \edge {x1} {x2};
        \edge {x2} {dots};
        \edge {dots} {xT};

        \edge {c} {x1};
        \edge {c} {x2};
        \edge {c} {xT};

        \edge {x1} {o1};
        \edge {x2} {o2};
        \edge {xT} {oT};

        \edge {x1} {o2};
        \edge {x1} {oT};
        \edge {x2} {oT};

        \draw[->, bend left=45] (x1) to (xT);
        \draw[->, bend left=45] (x2) to (xT);

        \plate {} {(x1)(x2)(xT)(o1)(o2)(oT)(dots)(odots)} {};
        \plate[dotted] {} {(x1)(x2)(dots)(xT)(llm)(x1p)(x2p)(xTp)} {};
        \plate[dotted] {} {(o1)(o2)(odots)(oT)(rm)(o1p)(o2p)(oTp)} {};

    \end{tikzpicture}
    \caption{
    State-space model for inference-time scaling. $c$ is a prompt, $x_1, \dots, x_T$ are LLM outputs, and $o_1, \dots, o_T$ are ``observed'' acceptances from a reward model. We estimate the latent states conditioned on $o_t = 1$ for all $t$.
    }
    \label{fig:plate}
    \vspace{-1em}
\end{figure}

\subsection{Particle Filtering for Estimating the Posterior}\label{sec:method-pf}
We now apply particle filtering (PF) to approximate the posterior over accepted sequences $x_{1:T}$ under the model in \eqref{eq:target}. Each particle represents a partial trajectory, and inference proceeds by iteratively generating, scoring, and resampling these particles based on their reward-induced likelihood.
At each time step $t$, PF maintains a population of $N$ weighted particles. The algorithm proceeds as follows:
\begin{itemize}
\item Initialization ($t=1$): Each particle is initialized by sampling the first token from the LLM:
$x_1^{(i)} \sim p_M(x_1 \mid c)$.
\item Propagation ($t > 1$): Each particle is extended by sampling the next token from the LLM conditioned on its history:
$x_t^{(i)} \sim p_M(x_t \mid c, x_{<t-1}^{(i)})$.
\item Weight Update: Using a reward model $\hat{r}$, each particle is assigned an unnormalized weight that reflects the likelihood of acceptance:
$w_t^{(i)} \propto w_{t-1}^{(i)} \cdot \hat{r}(c, x_{<t}^{(i)})$.
\item Resampling: To focus computation on promising trajectories, we sample a new population of particles from the current set using a softmax distribution over weights: $\mathbb{P}_t(j=i) = {\exp(w_t^{(i)})}/{\sum{i’=1}^{N} \exp(w_t^{(i’)})}$\label{eq:resampling-dist}.

\end{itemize}
The next generation of particles is formed by sampling indices $j_t^{(1)}, \dots, j_t^{(N)} \sim \mathbb{P}_t$ and retaining the corresponding sequences.

This procedure balances exploitation of high-reward partial generations with stochastic exploration, increasing the chances of recovering correct completions even when early steps are uncertain.

The algorithm can be implemented efficiently: both token sampling and reward computation can be parallelized across particles. With prefix caching, the total runtime is comparable to generating $N$ full completions independently. 

\paragraph{Final Answer Selection}
Particle filtering yields a weighted set of samples approximating the posterior, enabling principled answer selection strategies. While selecting the highest-weighted particle or using (weighted) majority voting better reflects the typical set, we adopt the standard practice of scoring all final outputs with an outcome reward model (ORM) and selecting the top-scoring one for fair comparison with prior work. Additionally, the posterior samples allow richer evaluation—for instance, estimating the expected correctness of the model under its own distribution, rather than relying solely on point estimates.
Notably, samples from \Algref{alg:pf} can be used to construct unbiased estimates of any expectation over \eqref{eq:target}.
In the context of math and reasoning, it guarantees the \emph{expected accuracy} is unbiased, which we formulate in Theorem~\ref{th:unbiasedness} (proof in Appendix~\ref{app-proof-unbiasedness}).

\begin{theorem}[Unbiasedness of Expected Accuracy]\label{th:unbiasedness}
Let $\{(w^{(i)}, x^{(i)})\}$ be weighted particles from \Algref{alg:pf} and $\mathrm{is\_correct}(x)$ is a function to check the correctness of response $x$.
We have
\begin{equation}\label{eq:unbiasedness}
    \mathbb{E} \left\{\sum_i\left[ w^{(i)} \; \mathrm{is\_correct}(x^{(i)}) \right] \right\} = \sum_{x}  \left[ \hat{p}_M(x_{1:T} \mid c, o_{1:T}=\mathbf{1}) \; \mathrm{is\_correct}(x^{(i)})  \right],
\end{equation}
where the expectation is over the randomness of the algorithm itself.
\end{theorem}

\paragraph{Reward Aggregation with PRMs}

To compute particle weights during generation, we aggregate per-step scores from the process reward model (PRM) $\hat{r}$. Our default uses a product of step-level rewards to align with the factorized likelihood structure, but alternative aggregation strategies (e.g., min, last-step, or model-based) may offer different trade-offs. We describe and compare these strategies in detail in \appref{app:aggregation} and report ablation results in \ref{results-rewardAggAblationStudy}.

\paragraph{Sampling v.s. deterministic search}
An alternative to our sampling-based approach is to treat inference-time scaling as an optimization problem under the approximate posterior \eqref{eq:target}, reducing to deterministic search methods like beam search or MCTS. However, these methods assume the reward model $\hat{r}$ is accurate at every step and prune aggressively based on early scores. In practice, PRMs are noisy and their preferences often shift during unrolling. As a result, deterministic search can irreversibly discard trajectories that may have low early scores but high posterior mass overall. Once pruned, such paths cannot be recovered.

In contrast, particle filtering maintains a distribution over partial sequences and uses stochastic resampling to balance exploration and exploitation. This allows recovery from early errors and robustness to reward noise. While beam search targets the mode of the approximate posterior-making it sensitive to local errors-PF samples from the typical set, smoothing over inconsistencies in $\hat{r}$. Unlike search heuristics, PF provides consistent, unbiased estimators under mild assumptions, and naturally captures multi-modal solutions—critical when multiple valid completions exist. We validate these advantages empirically in \secref{sec:res-ablation}.

\paragraph{Multiple iterations and parallel chains}\label{sec:method-pg-pt}

The PF approach to inference-time scaling can be used to define a MCMC kernel that enables two new types of scaling: multiple iterations of complete answers inspired by PG and parallel simulations inspired by parallel tempering. We detail the methodology and results for both extensions in sections \ref{sec:appendix-parallelGibbs} and \ref{sec:appendix-parallelTempering} in the appendix.

\section{Evaluation}

We evaluate our proposed methods in this section. We detail our experimental setup in \secref{sec:res-setup} and start with highlighted results comparing against closed-source models and competitive inference-time scaling methods with open-source models (\secref{sec:results-mathematical-reasoning-sections}). We then study how the main algorithm, particle filtering, scales with more computation and compare it with competitors (\secref{sec:res-scaling}). We further perform an extensive ablation study on key algorithmic choices like reward models, reward aggregation, final answer aggregation, and LLM temperatures (\secref{sec:res-ablation}). Finally, we study different allocations of the compute budget through iterative and parallel extensions (\secref{sec:res-alloc}).

\subsection{Setup}\label{sec:res-setup}

\paragraph{Models}
We consider two types of open-source small language models (SLMs) as our policy models for generating solutions. The first is general models, for which we evaluate Qwen2.5-1.5B-Instruct \citep{qwen_2_5}, Qwen2.5-7B-Instruct, Llama-3.2-1B-Instruct, and Llama-3.1-8B-Instruct \citep{llama3}. The second is math models, using Qwen2.5-Math-1.5B-Instruct and Qwen2.5-Math-7B-Instruct. These small models are well-suited for inference-time scaling, enabling efficient search of multiple trajectories.

\paragraph{Process Reward Models}
To guide our policy models, we utilized Qwen2.5-Math-PRM-7B \citep{prmlessons}, a 7B process reward model. 
We selected this model for its superior performance over other PRMs we tested, including Math-Shepherd-mistral-7b-prm \citep{wang2024mathshepherdverifyreinforcellms}, Llama3.1-8B-PRM-Deepseek-Data \citep{xiong2024rlhflowmath}, and EurusPRM-Stage2 \citep{primerm}. 
This result as an ablation study is provided in \secref{sec:res-ablation}, where we also study the different ways to aggregate step-level rewards from PRMs discussed in \secref{sec:method-pf}.


\paragraph{Baselines}
\begin{tightlist}
    \item Greedy: single greedy generation from the model, serving as the ``bottom-line'' performance.
    
    \item Self Consistency \citep{wang2023selfconsistencyimproveschainthought}:  simplest scaling method, majority voting across candidates

    \item BoN/WBoN \citep{brown2024largelanguage}: simple RM-based scaling method, scores outputs with outcome reward models
    
    \item Beam Search \citep{snell2024scalingllmtesttimecompute}: structured search procedure that incrementally builds sequences by keeping the top-$k$ highest-scoring partial completions at each generation step. 
    
    \item DVTS \citep{beeching2024scalingtesttimecompute}: a parallel extension of beam search that improves the exploration and performance. 
\end{tightlist}

\paragraph{Datasets}
We evaluate our methods and baselines across a wide variety of tasks that span difficulty level and domain to test basic and advanced problem-solving and reasoning. 
\begin{tightlist}
    \item \textbf{MATH500} \citep{math500}: 500  competition-level problems from various mathematical domains.
    \item \textbf{AIME 2024} \citep{ai_mo_validation_aime}: 30  high difficulty problems from the American Invitational Mathematics Examination (AIME I and II) 2024.
    \item \textbf{NumGLUE Task 2 (Chemistry)} \citep{mishra2022numglue}: 325 questions that test advanced reasoning across real-world tasks, mostly centered on the chemistry domain. 
    \item \textbf{FinanceBench} \citep{islam2023financebenchnewbenchmarkfinancial}: 150 open-book financial question answering tasks grounded in real-world financial documents and analysis.

\end{tightlist}

\paragraph{Parsing and scoring}
Details on parsing and scoring functions across datasets in the Appendix \ref{app:eval} 

\subsection{Results on Mathematical Reasoning Datasets}\label{sec:results-mathematical-reasoning-sections}

\begin{figure}[h!]
  \centering

  \begin{subfigure}[t]{0.49\textwidth}
    \includegraphics[width=\linewidth]{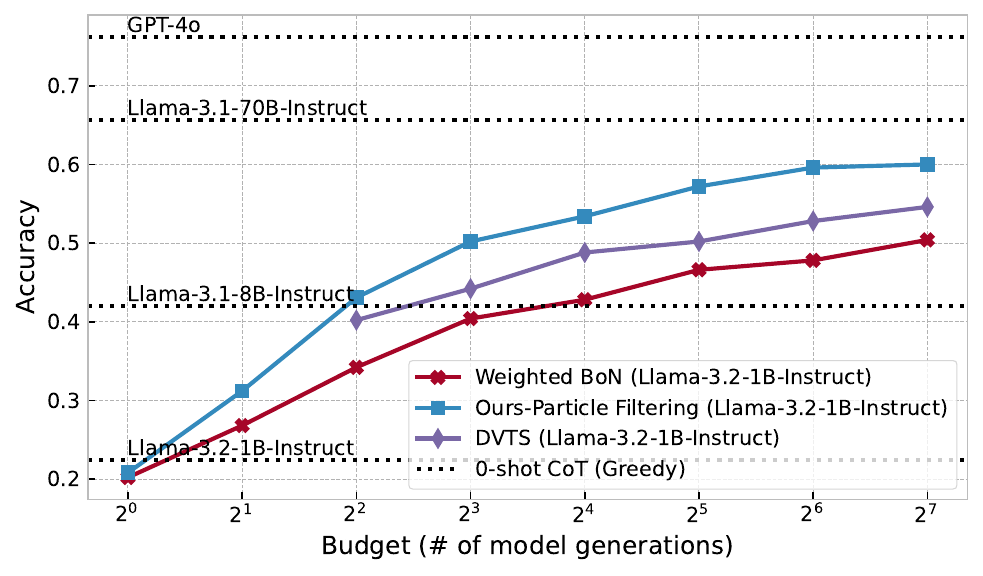}
    \caption{Llama-3.2-1B-Instruct}
  \end{subfigure}%
  \hspace{0.5mm}%
  \begin{subfigure}[t]{0.49\textwidth}
    \includegraphics[width=\linewidth]{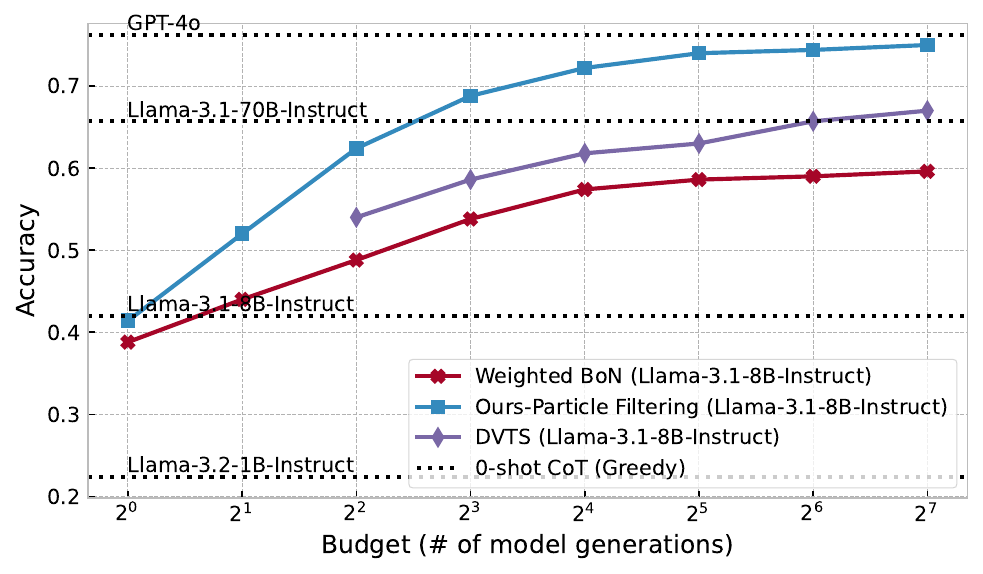}
    \caption{Llama-3.1-8B-Instruct}
  \end{subfigure}

  \vspace{1mm}

  \begin{subfigure}[t]{0.49\textwidth}
    \includegraphics[width=\linewidth]{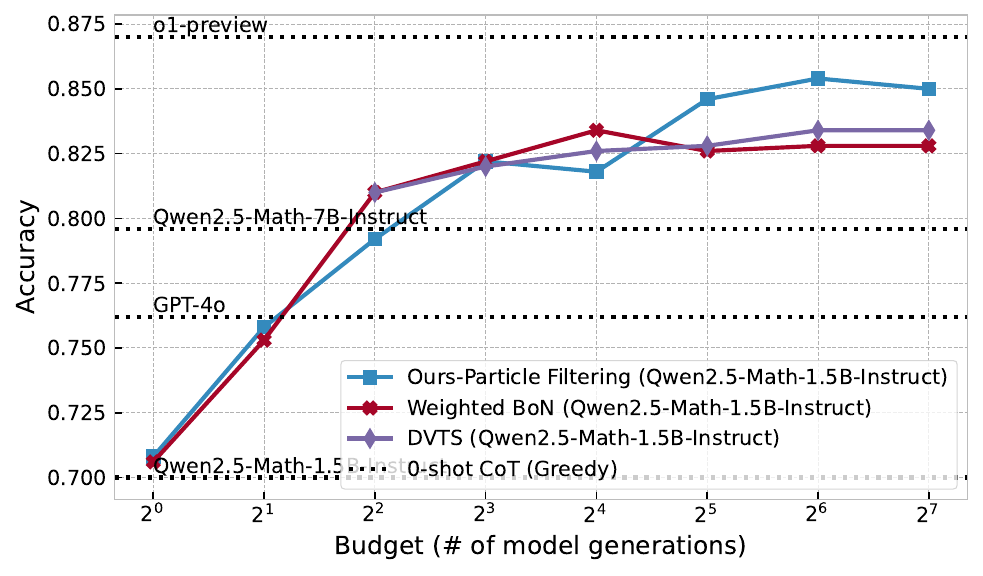}
    \caption{Qwen2.5-Math-1.5B-Instruct}
  \end{subfigure}%
  \hspace{0.5mm}%
  \begin{subfigure}[t]{0.49\textwidth}
    \includegraphics[width=\linewidth]{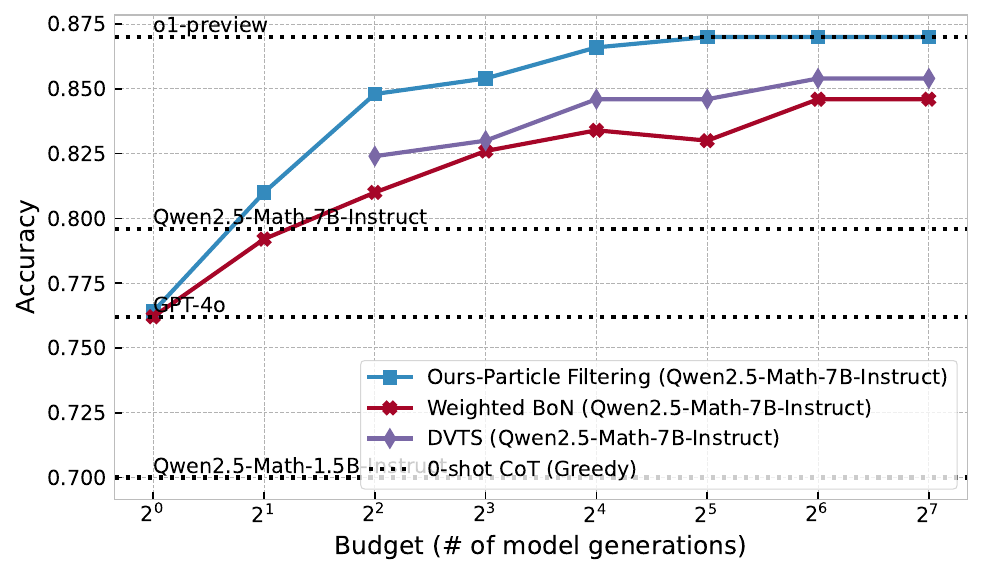}
    \caption{Qwen2.5-Math-7B-Instruct}
  \end{subfigure}

  \caption{Accuracy vs. Generation Budget across models using different inference-time strategies.}
  \label{fig:scaling_plot}
\end{figure}


We present our main results in Table~\ref{tab:maintable-performance}, comparing Particle Filtering (PF) with a suite of strong inference-time scaling baselines on two challenging mathematical reasoning tasks: MATH500 and AIME 2024. All inference-time scaling methods are evaluated under a fixed compute budget of 32 generations per instance, using Qwen2.5-Math-PRM-7B as the reward model. Specifically, it serves as the PRM for PF, Beam Search, and DVTS, and as the ORM in WBoN.

\begin{itemize}

    \item \textit{PF consistently achieves the best performance across all model sizes.} 
    Among all inference-time scaling methods, PF delivers the highest accuracy on both benchmarks, often outperforming alternatives by a significant margin.
    \item \textit{PF unlocks competitive performance even for small models.} 
    For instance, Qwen2.5-1.5B-Instruct, when scaled using PF, surpasses the much larger GPT-4o on both MATH500 and AIME 2024. This showcases the ability of inference-time compute to significantly improve performance without increasing model size.
    \item \textit{PF on Qwen2.5-Math-7B-Instruct outperforms o1-preview on MATH500.} 
    Scaling Qwen2.5-Math-7B with PF results in a new state-of-the-art among open models: 87.7\% on MATH500 and 10/30 on AIME. This surpasses the o1-preview model and highlights the potential of inference-time scaling to close the gap with—or even exceed—the performance of proprietary frontier LLMs using smaller, open models.

\end{itemize}
 
For results on additional model families and broader ablations, see Appendix~\ref{app-moreResultsMoreModels}.

\begin{table}[]
\centering
\begin{tabular}{lccc}
\hline
\textbf{Model}              & \multicolumn{1}{c|}{\textbf{Method}}                                   & \textbf{MATH500}         & \textbf{AIME 2024}       \\ \hline
\multicolumn{4}{l}{\textbf{Closed Source LLMs}}                                                                                                            \\
GPT-4o                      & \multicolumn{1}{r|}{--}                                                & 76.2            & 4/30                     \\
o1-preview                  & \multicolumn{1}{r|}{--}                                                & \textbf{87.0}            & \textbf{12/30}                    \\
Claude 3.5 Sonnet           & \multicolumn{1}{r|}{--}                                                & 78.2            & 5/30                     \\ \hline
\multicolumn{4}{l}{\textbf{Open Source LLMs}}                                                                                                              \\
Llama-3.1 70B Instruct      & \multicolumn{1}{r|}{--}                                                & 65.6            & 5/30                     \\
Qwen-2.5 Math 72B Instruct  & \multicolumn{1}{r|}{--}                                                & 82.0            & 9/30                     \\ \hline
\multicolumn{4}{l}{\textbf{Open Source General SLMs}}                                                                                                      \\
Qwen-2.5 1.5B Instruct      & \multicolumn{1}{r|}{Greedy}                                            & 54.4            & 1/30                     \\
                            & \multicolumn{1}{r|}{Self Consistency}                                  & 61.0            & 2/30                     \\
                            & \multicolumn{1}{r|}{BoN}                                               & 67.8            & 1/30                     \\
                            & \multicolumn{1}{r|}{WBoN}                                              & 69.2            & 2/30                     \\
                            & \multicolumn{1}{r|}{Beam Search}                                       & 76.2            & 5/30                     \\
                            & \multicolumn{1}{r|}{DVTS}                                              & 76.6             & 4/30                     \\
\rowcolor[HTML]{C1DEB8} 
                            & \multicolumn{1}{r|}{\cellcolor[HTML]{C1DEB8}\textbf{Particle Filtering (Ours)}} & \textbf{79.3}            & \textbf{6/30}                     \\ \hline
\multicolumn{4}{l}{\textbf{Open Source Math SLMs}}                                                                                                         \\
Qwen-2.5 Math 1.5B Instruct & \multicolumn{1}{r|}{Greedy}                                            & 70.0            & 3/30                     \\
                            & \multicolumn{1}{r|}{Self Consistency}                                  & 79.6  & 6/30 \\
                            & \multicolumn{1}{r|}{BoN}                                               & 81.8            & 4/30                     \\
                            & \multicolumn{1}{r|}{WBoN}                                              & 82.6            & 4/30                     \\
                            & \multicolumn{1}{r|}{Beam Search}                                       & 83.0  & 4/30 \\
                            & \multicolumn{1}{r|}{DVTS}                                              & 82.8            & 5/30                     \\
\rowcolor[HTML]{C1DEB8} 
                            & \multicolumn{1}{r|}{\cellcolor[HTML]{C1DEB8}\textbf{Particle Filtering (Ours)}} & \textbf{84.6 }  & \textbf{7/30}            \\ \hline
Qwen-2.5 Math 7B Instruct   & \multicolumn{1}{r|}{Greedy}                                            & 79.6            & 5/30                     \\
                            & \multicolumn{1}{r|}{Self Consistency}                                  & 84.0  & 4/30                     \\
                            & \multicolumn{1}{r|}{BoN}                                               & 82.6            & 5/30                     \\
                            & \multicolumn{1}{r|}{WBoN}                                              & 83.0            & 5/30                     \\
                            & \multicolumn{1}{r|}{Beam Search}                                       & 86.9  & 7/30 \\
                            & \multicolumn{1}{r|}{DVTS}                                              & 84.6            & 6/30                     \\
\rowcolor[HTML]{C1DEB8} 
                            & \multicolumn{1}{r|}{\cellcolor[HTML]{C1DEB8}\textbf{Particle Filtering (Ours)}} & \textbf{87.7 }  & \textbf{10/30}           
\end{tabular}
\vspace{.5em}
\caption{Results of various LLMs on MATH500 and AIME 2024, highlighting particle filtering performance. All methods used a compute budget of 32 generations with Qwen2.5-Math-PRM-7B as the reward model. Notably, Qwen2.5-Math-7B, with just 32 particles, matches o1-preview on MATH500, demonstrating PF's effectiveness.}
\vspace{-2em}
\label{tab:maintable-performance}
\end{table}

\subsection{Results on Generalized Reasoning Datasets}\label{sec:results-nonmath-reasoning-section}
To evaluate whether our inference-time scaling method generalizes beyond mathematical reasoning, we test Particle Filtering on two diverse instruction-following benchmarks: \textbf{FinanceBench} \citep{islam2023financebenchnewbenchmarkfinancial}, which evaluates financial QA over real-world documents, and \textbf{NumGLUE Task 2 (Chemistry)} \citep{mishra2022numglue}, which targets numerical reasoning in scientific contexts.

\begin{wraptable}{R}{0.63\textwidth}
\centering
\vspace{0em} 
\begin{tabular}{c|c|c}
\textbf{Method}                                                      & \textbf{FinanceBench} & \textbf{\begin{tabular}[c]{@{}c@{}}NumGLUE Task 2\\ (Chemistry)\end{tabular}} \\ \hline
Greedy                                                               & 62.67                 & 71.69                                                                         \\
BoN                                                                  & 68.00                 & 80.92                                                                         \\
Self Consistency                                                     & 68.67                 & 79.32                                                                         \\
Beam Search                                                          & 67.33                 & 80.47                                                                         \\
\rowcolor[HTML]{C1DEB8} 
\begin{tabular}[c]{@{}c@{}}Particle \\ Filtering (Ours)\end{tabular} & 70.33                 & 84.22                                                                        
\end{tabular}
\caption{Results on Non-Math Datasets}
\vspace{-1em}
\label{tab:non-math-dataset-results-table}
\end{wraptable}

As shown in Table~\ref{tab:non-math-dataset-results-table}, Particle Filtering consistently outperforms all other inference-time scaling baselines, achieving the highest accuracy on both datasets. This shows that our method is effective not only for mathematical reasoning but also for broader instruction-following and domain-specific tasks. We use Llama 3.1-8B-Instruct as the policy model and Qwen2.5-Math-PRM-7B as the reward model, with 8 particles for FinanceBench and 32 for NumGLUE.

We note that although we use Qwen2.5-Math-PRM-7B - a reward model trained primarily for mathematical process evaluation - it performs surprisingly well as a reward model on these non-math domains. We hypothesize that such RMs implicitly learn broader reasoning evaluation capabilities during their training, not limited strictly to mathematical content. We leave deeper exploration of this hypothesis and the development of domain-specific or generalized PRMs for future work.

\subsection{Scaling with inference-time compute}\label{sec:res-scaling}
We now zoom in on how PF scales with inference-time compute.
\Figref{fig:scaling_plot} shows the change of performance (in terms of accuracy) with an increasing computation budget ($N=1, 2, 4, 8, 16, 32, 64, 128$) across Math SLMs and Non-Math SLMs. 
As we can see, PF scales 4--16x faster than the next best competitor DVTS, 
e.g.~DVTS requires a budget of 32 to reach the same performance of PF with a budget of 8 with Llama-3.2-1B-Instruct and requires a budget of 128 to reach the performance of PF with a budget of 8 with LLama-3.1-8B-Instruct.

\subsection{Ablation study}\label{sec:res-ablation}


\begin{wrapfigure}{r}{0.5\textwidth}
    \centering
    \vspace{-1.75em} 
    \begin{subfigure}[h]{0.48\textwidth}
        \centering
        \includegraphics[width=\linewidth]{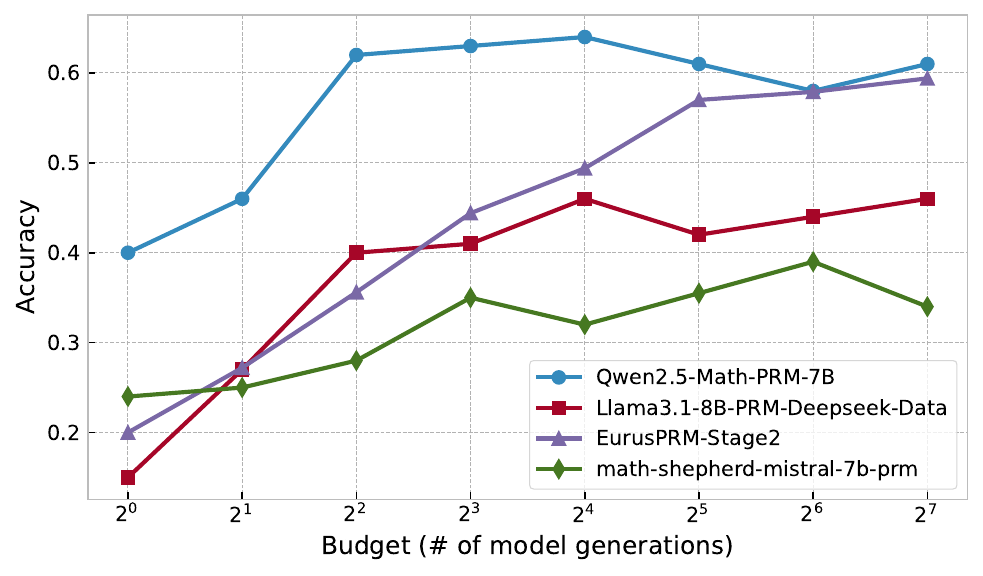}
        \caption{Results of ablation comparing the performance of PF across PRMs.}
        \label{fig:rm_comparison}
    \end{subfigure}
    
    \begin{subfigure}[h]{0.48\textwidth}
        \centering
        \includegraphics[width=\linewidth]{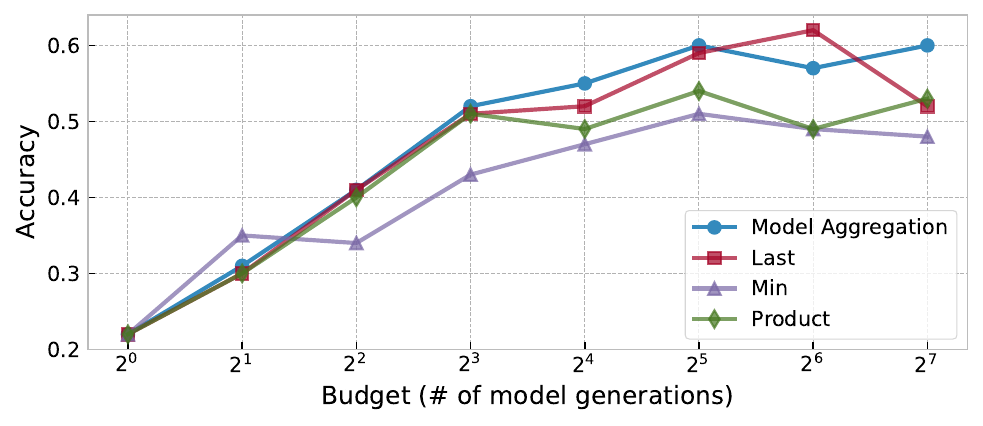}
        \caption{Effect of different aggregation strategies for Qwen2.5-Math-PRM-7B.}
        \label{fig:aggregation_comparison}
    \end{subfigure}

    \begin{subfigure}[h]{0.48\textwidth}
        \centering
        \includegraphics[width=\linewidth]{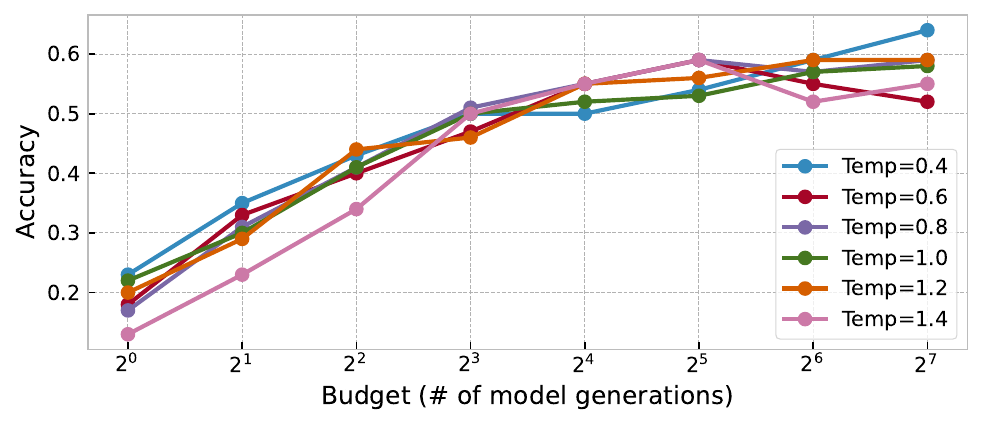}
        \caption{Results of ablation comparing the effect of temperature across different particle budget.}
        \label{fig:model_temp_sweep_plot}
    \end{subfigure}

    \vspace{-5em}
\end{wrapfigure}

\paragraph{Performance of different PRMs}
Figure~\ref{fig:rm_comparison} shows an ablation on 100 MATH500 questions, comparing our method’s accuracy across reward functions as the number of particles increases.


Qwen2.5-Math-PRM-7B consistently outperforms other models, making it the natural choice for our main results. Interestingly, while EurusPRM-Stage2 performs poorly with smaller budgets, it improves and eventually matches Qwen2.5-Math-PRM-7B at higher budgets.

\paragraph{Reward aggregation within PRMs} \label{results-rewardAggAblationStudy}

As mentioned in \secref{sec:method-pf} / Appendix \ref{app:aggregation} and reported by prior works \citep{zhang2025lessonsdeveloping}, there are multiple ways to use PRMs to calculate reward scores, which can significantly impact final performance. \Figref{fig:aggregation_comparison} studies three existing methods for combining PRM scores—using the \textit{last} reward, the \textit{minimum} reward, and the \textit{product} of all rewards. We also study ``Model Aggregation," where the PRM is used as an ORM with partial answers.


As we can see, using Model Aggregation---in essence, feeding into a PRM the entire partial answer alongside the question - scales the best with an increasing budget.\looseness=-1

\paragraph{Controlling the state transition---temperatures in LLM generation}

We investigate the effect of different LM sampling temperatures on the scaling of our method across varying numbers of particles. The results of our ablation study on a 100-question subset of MATH questions are shown in Figure \ref{fig:model_temp_sweep_plot}.


Our findings show that the common LLM temperature range of 0.4--1.0 performs well, with minimal accuracy variation across budgets. Following \citep{beeching2024scalingtesttimecompute}, we use temperature 0.8 for all experiments.


\paragraph{Budget allocation over iterations and parallelism}\label{sec:res-alloc}
The multi-iteration and parallel-chain extensions introduced in \secref{sec:method-pg-pt} / Appendix \ref{sec:appendix-multipleIterationsParallelChains} provide two more axes to spend computation beyond the number of particles. We explore how different budget allocations affect performance in the appendix.

\section{Related Work}

\textbf{Process reward models} (PRMs) aim to provide more granular feedback by evaluating intermediate steps rather than only final outputs.
They are trained via process supervision, a training approach where models receive feedback on each intermediate step of their reasoning process rather than only on the final outcome.
\citep{lightman2023letsverify} propose a step-by-step verification approach to PRMs, improving the reliability of reinforcement learning.
DeepSeek PRM \cite{wang2024mathshepherdverifyreinforcellms} uses Mistral to annotate training data for PRMs .
\citep{zhang2025lessonsdeveloping} introduces Qwen-PRM, which combines both Monte Carlo estimation and model/human annotation approach to prepare training data for a PRM. 
PRIME~\citep{cui2024process} proposes to train an outcome reward model (ORM) using an implicit reward objective. The paper shows that implicit reward objective directly learns a Q-function that provides rewards for each token, which can be leveraged to create process-level reward signal. This process eliminates the need for any process labels, and reaches competitive performance on PRM benchmarks.

\textbf{Inference-time scaling} is a key training-free strategy for enhancing LLM performance. \citep{brown2024largelanguage} investigates best-of-N (BoN) decoding, showing quality gains via selective refinement. \citep{snell2024scalingllm} analyzes how scaling compute improves inference efficiency from a compute-optimality view. While not implementing full Monte Carlo tree search (MCTS), \citep{zhou2024languageagenttreesearch} explores a tree-search-inspired approach within language models. \citep{guan2025rstarmathsmall} introduces rSTAR, which combines MCTS for data generation and training to improve mathematical reasoning. \citep{beeching2024scalingtesttimecompute} discusses beam search and dynamic variable-time search (DVTS) as inference-time scaling methods for open-source LLMs. DVTS runs multiple subtrees in parallel to avoid all leaves getting stuck in local minima.

\textbf{Particle-based Monte Carlo methods} are powerful tools for probabilistic inference. Sequential Monte Carlo \citep{sequentialmonte} or particle filtering \citep{nonlinearfiltering} has been a classical way to approximate complex posterior distributions over state-space models. Particle Gibbs (PG) sampling \citep{andrieu2010particlemarkov} extends these approaches by integrating MCMC techniques for improved inference. \citep{lew2023sequentialmontecarlosteering} and \citep{loula2025syntactic} use token-based SMC within probabilistic programs to steer LLMs, while \citep{grand2025selfsteeringlanguagemodels} apply token-based SMC for self-constrained generation. \cite{zhao2024probabilisticinferencelanguagemodels} and \cite{feng2024stepbystepreasoningmathproblems} introduce Twisted SMC for inference in language models. We note that our method differs from \cite{feng2024stepbystepreasoningmathproblems} in several key ways.
First, TSMC relies on a ground-truth verifier and requires joint training of a generator and value function, whereas our approach uses an \textit{off-the-shelf} generator and a noisy pretrained reward model (PRM), requiring \textit{no additional training}. Second, our method generalizes to domains lacking ground-truth verifiers, such as instruction and broader language tasks. Finally, we demonstrate significantly stronger empirical results. The authors of TSMC did not release code, and their method requires additional training. We therefore compare our method—using the same generator model (DeepSeekMath7B) and dataset (MATH500) as in their results table—and find it achieves 75.4\% accuracy with 128 samples, outperforming TSMC by 14.6 points using fewer than half the samples and no fine-tuning.

\textbf{Decision making with uncertainty}
The way of representing uncertainty using softmax is often referred as Boltzmann exploration in the multi-armed bandit (MAB) literature \citep{kuleshov2014algorithms}.
While formulating it as a MAB problem allows it to use a scheduling on the softmax temperature and to derive regret bounds \citep{cesa2017boltzmann}, we no longer have the same unbiasedness from the particle filtering / SMC formulation.

\section{Conclusion}
In this paper, we introduce a particle filtering algorithm for inference-time scaling. To address the limitations of deterministic inference-time scaling—namely, early pruning from imperfect reward models—we adapt particle-based Monte Carlo methods to maintain a diverse population of candidate sequences and balance exploration and exploitation. This probabilistic framing enables more resilient generation and opens a principled path for integrating uncertainty into LLM inference. Our evaluation shows these algorithms consistently outperform search-based approaches by a significant margin.

However, inference-time scaling comes with computational challenges. Hosting and running a reward model often introduces high latency, making the process more resource-intensive. For smaller models, prompt engineering is often required to ensure outputs adhere to the desired format. Finally, hyperparameters like budget are problem-dependent and may require tuning across domains.

We hope that the formal connection of inference scaling to probabilistic modeling established in this work will lead to systematic solutions for current limitations of these methods and pave the way for bringing advanced probabilistic inference algorithms into LLM inference-time scaling in future work.

\bibliography{main}
\bibliographystyle{plain}


\newpage
\appendix

\section{Aggregation Strategies for PRM Scores}
\label{app:aggregation}

To compute particle weights during generation, we aggregate per-step scores from the process reward model (PRM) $\hat{r}$. Our default uses a product of step-level rewards to align with the factorized likelihood structure, but alternative aggregation strategies (e.g., min, last-step, or model-based) may offer different trade-offs. 

The weight update step in particle filtering depends on how rewards are assigned to partial trajectories using a preference or reward model (PRM) $\hat{r}$. Since PRMs often provide per-step scores, aggregating them into a single weight requires a strategy that balances theoretical correctness and practical utility.

We consider the following four aggregation approaches:

\begin{itemize}
    \item Product ($\mathrm{prod}$): Computes the product of step-level rewards across all tokens. This aligns directly with the factorized likelihood structure used in the PF objective (\eqref{eq:target}), enabling online weight updates as generation proceeds.

    \item Minimum ($\mathrm{min}$): Takes the minimum reward seen so far. This penalizes trajectories for weak intermediate steps, which may help in discouraging risky completions. However, it prevents online updates because the entire prefix must be scored to determine the weight.
    
    \item Last-step ($\mathrm{last}$): Uses only the most recent step’s reward. Although not aligned with a likelihood-based interpretation, this method is computationally efficient and reflects the scoring mode used in \citep{beeching2024scalingtesttimecompute}.
    
    \item Model-based aggregation: Instead of relying on step-wise rewards, this method repurposes the PRM in a black-box fashion to assign a single scalar score to the full partial trajectory. This helps smooth over noisy token-level scores and can be more stable, especially when PRMs are inconsistent across steps. The model receives the prompt and prefix and returns a scalar reward.
\end{itemize}

\appref{app:example} shows how the input format for this black-box mode differs from standard per-step PRM usage. We compare all strategies empirically in \ref{results-rewardAggAblationStudy} and find that the optimal choice varies with the PRM’s training and evaluation objective.

\section{Multiple Iterations and Parallel Chains}\label{sec:appendix-multipleIterationsParallelChains}

Here, we explore how different ways to allocate budgets change the performance in the appendix. 
Specifically, we study for a fixed budget $N \times T \times M$, how the combination of $N,T,M$ can yield the best performance, where $N$ is the number of particles, $T$ is the number of iterations, and $M$ is the number of parallelism.

\subsection{Particle Gibbs}\label{sec:appendix-parallelGibbs}
\textbf{Particle Gibbs} is a type of MCMC algorithm that uses PF as a transition kernel \citep{andrieu2010particlemarkov}. 
Specifically, at each iteration, PG samples a new set of particles  using PF with a reference particle from the previous iteration. 
This integration combines the efficiency of PF with the theoretical guarantees of MCMC, making PG suitable for high-dimensional or challenging posterior distributions.
The adaption of PG to inference-time scaling is essentially a multi-iteration extension of the PF algorithm presented, which works as follows:
For each iteration, we run a modified PF step with an additional sampling step to sample 1 reference particle according to the equation in the resampling step of Section \ref{eq:resampling-dist}.
For any PF step that is not the initial step, the PF is executed with a reference particle: This reference particle is never replaced during the resampling step, but its partial trajectory can still be forked during resampling.
We detail the PG version of inference-time scaling in \algref{alg:pg} of Appendix \ref{appendix-pt-pg-algs}.
Note that typically, a reasonably large number of particles is needed to show the benefits of multiple iterations, which we also confirm in our results in \secref{sec:res-ablation}.

\paragraph{Allocating budget between $N$ and $T$}

\begin{figure}[h]
    \centering
    \includegraphics[width=0.77\linewidth]{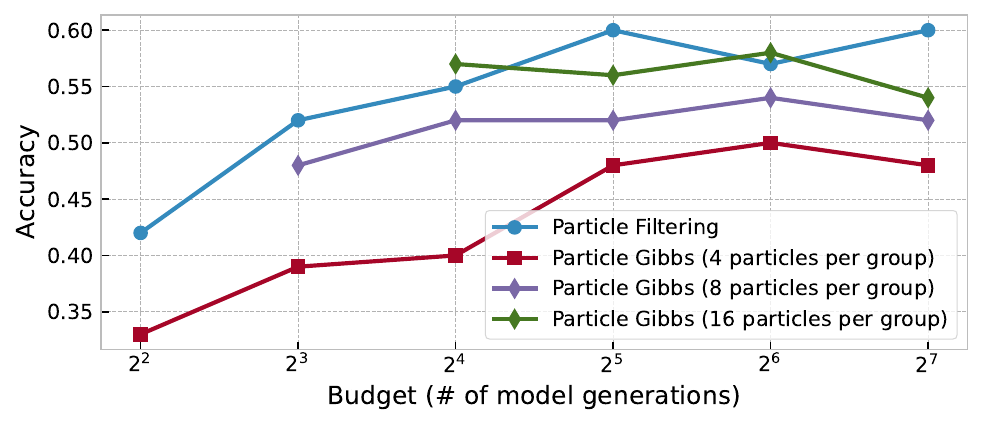}
    \caption{Comparison of PF and Particle Gibbs with different numbers of iterations, evaluated on a 100-question subset of the MATH-500 dataset using Llama-3.2-1B-Instruct as the policy model.}
    \label{fig:pg_analysis}
\end{figure}

Figure \ref{fig:pg_analysis} shows results of Llama-3.2 1B model when configured with various test-time compute budget allocations. Although the plot shows that various Particle Gibbs configurations do not have a marked benefit over an equivalently budgeted particle filtering run, a PG experiment with 16 particles and 4 iterations powered by a Qwen 2.5 7B Math Instruct policy model achieved a 87.2\% accuracy on MATH500, beating o1 performance. Configurations with larger $N$ values typically do better than equivalently budgeted runs with less particles.

\subsection{Parallel Tempering}\label{sec:appendix-parallelTempering}
\paragraph{Parallel tempering}
In parallel tempering (aka replica exchange MCMC sampling), multiple MCMC chains run in parallel at different temperatures and swap the states to allow better exploration.
The key idea is that the chain running in high temperature can explore better, e.g.~traversing between different modes of the target, and the swap makes it possible to let the low temperature chain exploit the new region found by the other chain.
We detail the complete parallel tempering version of inference-time scaling in the Algorithms section of the Appendix below \algref{alg:pt} of \ref{appendix-pt-pg-algs}, while we only explore a special case of it (multiple chains with single iteration) in our experiments.

\paragraph{Allocating budget between $N$ and $M$}

\begin{figure}[h]
    \centering
    \includegraphics[width=0.77\columnwidth]{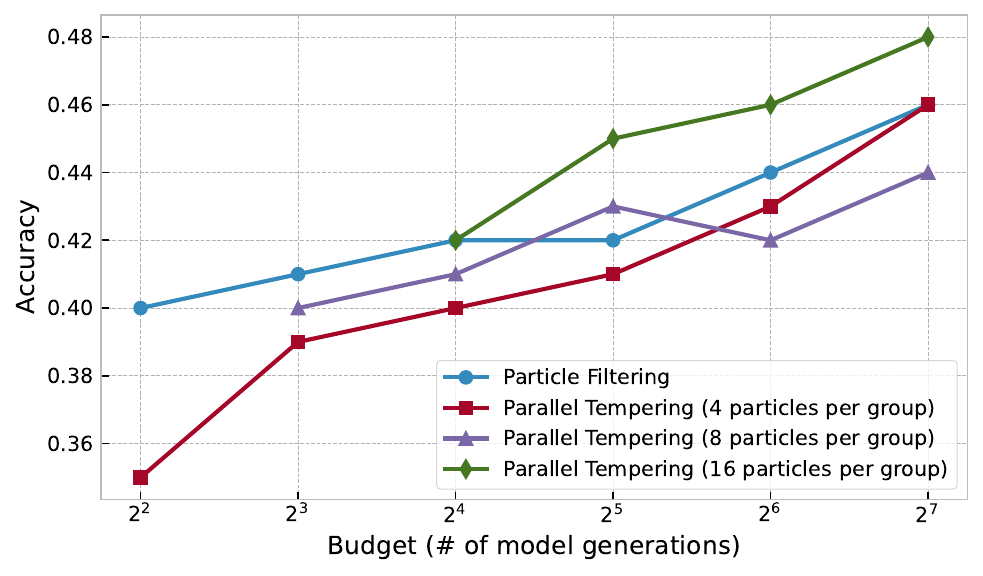}
    \caption{
    Comparison of PF and PT with different particle group sizes, evaluated on a 100-question subset of the MATH500 dataset using Llama-3.2-1B-Instruct as the policy model.}
    \label{fig:pt_comparison}
    \vspace{-1.5em}
\end{figure}

\Figref{fig:pt_comparison} shows PF and 3 PT configurations over a set of increasing numbers of budgets.
First, as we can see, for any fixed $N$, increasing $M$ also improves the performance.
This may be helpful when combining batch generation with distributed computing.
Second, PT with $N=16$ has a better overall scaling than PF.
This indicates that there is some optimal budget allocation over parallel chains that can further improve the overall performance of our main results.

We leave the exploration over the optimal configuration of $N,T,M$ jointly as a future work.

\section{Proof of Theorem \ref{th:unbiasedness}}\label{app-proof-unbiasedness}

\begin{theorem}[Unbiasedness of Expected Accuracy]\label{th:unbiasedness}
Let $\{(w^{(i)}, x^{(i)})\}$ be weighted particles from \Algref{alg:pf} and $\mathrm{is\_correct}(x)$ is a function to check the correctness of response $x$.
We have
$$
    \mathbb{E} \left\{\sum_i\left[ w^{(i)} \; \mathrm{is\_correct}(x^{(i)}) \right] \right\} = \sum_{x}  \left[ \hat{p}_M(x_{1:T} \mid c, o_{1:T}=\mathbf{1}) \; \mathrm{is\_correct}(x^{(i)})  \right],
$$
where the expectation is over the randomness of the algorithm itself.
\end{theorem}

\begin{proof}
This is a direct result of applying the unbiasedness property of particle filtering on a well-defined expectation $p$ $\mathbb{E}_{x \sim p}\{f(x)\}$ of any function $f$ over a distribution $p$: the Monte Carlo estimate using weighted samples from particle filtering is an unbiased estimate of this expectation.
As $\mathrm{is\_correct}(\cdot)$ is a binary function, the expectation of the estimate is finite thus well-defined and therefore the unbiasedness of accuracy holds.
\end{proof}

\section{Side by Side Comparison of Particle Filtering vs Beam Search}\label{appendix-pt-pg-algs}

\begin{figure}[h]
    \centering
    \begin{subfigure}[t]{0.48\textwidth}
        \centering
        \resizebox{\linewidth}{!}{
            \begin{tikzpicture}
    \tikzstyle{problem} = [rectangle, draw=black, fill=blue!20, rounded corners, minimum width=2cm, minimum height=0.5cm]
    \tikzstyle{node} = [circle, draw=black, minimum size=8mm, inner sep=0mm]
    \tikzstyle{selected} = [diamond, draw=black, fill=green!30, minimum size=6mm]
    \tikzstyle{unselected} = [diamond, draw=black, fill=red!30, minimum size=6mm]
    \tikzstyle{final} = [circle, draw=black, minimum size=6mm]
    \tikzstyle{bar} = [rectangle, draw=black, fill=blue!50]
    
    \node[problem] (root) {math problem};
    
    \node[selected] (n1) [below left=1.0cm and 0.5cm of root] {\tiny +2.1};
    \node[unselected] (n2) [below left=1.0cm and -0.75cm of root] {\tiny -1.2};
    \node[selected] (n3) [below right=1.0cm and -0.75cm of root] {\tiny +1.3};
    \node[selected] (n4) [below right=1.0cm and 0.5cm of root] {\tiny +0.1};
    
    \node[final] (f1) [below=0.65cm of n1] {};
    \node[final] (f2) [below=0.65cm of n2] {};
    \node[final] (f3) [below=0.65cm of n3] {};
    \node[final] (f4) [below=0.65cm of n4] {};
    
    \draw (root) -- (n1);
    \draw (root) -- (n2);
    \draw (root) -- (n3);
    \draw (root) -- (n4);
    
    \draw[dashed] (n1) -- (f1);
    \draw[dashed] (n1) -- (f2);
    \draw[dashed] (n3) -- (f3);
    \draw[dashed] (n4) -- (f4);
    
    \node[above right=0.1cm and 0.5cm of n4] {\textcolor{blue}{softmax}};
    \node[right=0.5cm of f4] {\textcolor{blue}{sample $N$: $1, 1, 3, 4$}};

    \node[below right=-0.1cm and 0.5cm of n4] {\textcolor{blue}{distribution:}};
    \node[below right=0.35cm and 2.45cm of n4] (b1) {};
    \node[right=0.01cm of b1] (b2) {};
    \node[right=0.01cm of b2] (b3) {};
    \node[right=0.01cm of b3] (b4) {};
    
    \draw[bar] ($(b1) + (-0.1,0)$) rectangle ($(b1) + (0.1,+0.617)$);
    \draw[bar] ($(b2) + (-0.1,0)$) rectangle ($(b2) + (0.1,+0.023)$);
    \draw[bar] ($(b3) + (-0.1,0)$) rectangle ($(b3) + (0.1,+0.277)$);
    \draw[bar] ($(b4) + (-0.1,0)$) rectangle ($(b4) + (0.1,+0.083)$);
\end{tikzpicture}
        }
        \caption{Particle filtering uses the rewards to produce a softmax distribution and does stochastic expansion of $N$ based sampling.}
        \label{fig:pf}
    \end{subfigure}
    \hfill
    \begin{subfigure}[t]{0.48\textwidth}
        \centering
        \resizebox{\linewidth}{!}{
            \begin{tikzpicture}
    \tikzstyle{problem} = [rectangle, draw=black, fill=blue!20, rounded corners, minimum width=2cm, minimum height=0.5cm]
    \tikzstyle{node} = [circle, draw=black, minimum size=8mm, inner sep=0mm]
    \tikzstyle{selected} = [diamond, draw=black, fill=green!30, minimum size=6mm]
    \tikzstyle{unselected} = [diamond, draw=black, fill=red!30, minimum size=6mm]
    \tikzstyle{final} = [circle, draw=black, minimum size=6mm]
    
    \node[problem] (root) {math problem};
    
    \node[selected] (n1) [below left=1cm and 0.5cm of root] {\tiny +2.1};
    \node[unselected] (n2) [below left=1cm and -0.75cm of root] {\tiny -1.2};
    \node[selected] (n3) [below right=1cm and -0.75cm of root] {\tiny +1.3};
    \node[unselected] (n4) [below right=1cm and 0.5cm of root] {\tiny +0.1};
    
    \node[final] (f1) [below=0.65cm of n1] {};
    \node[final] (f2) [below=0.65cm of n2] {};
    \node[final] (f3) [below=0.65cm of n3] {};
    \node[final] (f4) [below=0.65cm of n4] {};
    
    \draw (root) -- (n1);
    \draw (root) -- (n2);
    \draw (root) -- (n3);
    \draw (root) -- (n4);
    
    \draw (n1) -- (f1);
    \draw (n1) -- (f2);
    \draw (n3) -- (f3);
    \draw (n3) -- (f4);
    
    \node[above right=0.1cm and 0.5cm of n4] {\textcolor{blue}{top-${N \over M}$}};
    \node[below right=-0.1cm and 0.5cm of n4] {\textcolor{blue}{selected: $1, 3$}};
    \node[right=0.5cm of f4] {\textcolor{blue}{expand $M$}};
\end{tikzpicture}
        }
        \caption{Beam search treats the rewards as exact and performs deterministic expansion based on beam size $N$ and beam width $M$.}
        \label{fig:bs}
    \end{subfigure}
    \caption{A side-by-side comparison between particle filtering and its closest search-based counterpart, beam search. Compared with beam search in \figref{fig:bs} where the selection and expansion is deterministic (implicitly assumes the rewards are correct), particle filtering in \figref{fig:pf} trusts the rewards with uncertainty and propagates the expansion via sampling.}
    \label{fig:illustration}
\end{figure}

\section{Algorithms}

\setlength{\textfloatsep}{4pt}
\begin{algorithm}[h]
\caption{Particle Filtering for Inference-Time Scaling}\label{alg:pf}
\begin{algorithmic}
\State \textbf{Input}: the number of particles $N$, a reward model $\hat{r}$, a LLM $p_M$ and the prompt $c$
\State Initialize $N$ particles $\{x_1^{(i)} \sim p_M(\cdot \mid c)\}_{i=1}^N$
\State $t \gets 1$
\While{not all particles stop} 
    \State Update rewards $\mathbf{w} = [\hat{r}(x_{1:t}^{(1)}), \dots, \hat{r}(x_{1:t}^{(N)})]$
    \State Compute softmax distribution $\theta = \mathrm{softmax}(\mathbf{w})$
    \State Sample indices $\{j_t^{(i)}\}_{i=1}^N  \sim \mathbb{P}_t(j=i) = \theta_i$
    \State Update the set of particles as $\{x_{1:t}^{(j_t^{(i)})}\}_{i=1}^N$
    \State Transition $\{x_{t+1}^{(i)} \sim p_M(\cdot \mid c, x_{1:t}^{(i)})\}_{i=1}^N$
    \State $t \gets t + 1$
\EndWhile
\State \textbf{Return}: the set of weighted particles in the end
\end{algorithmic}
\end{algorithm}
\setlength{\textfloatsep}{20pt}

\newpage

\begin{algorithm}[h]
\caption{Particle Gibbs for Inference-Time Scaling}\label{alg:pg}
\begin{algorithmic}
\State \textbf{Input}: same as \algref{alg:pf} with the number of Gibbs iterations $T$
\State Run \algref{alg:pf} to get a set of particles $\{x_{1:t}^{(i)}\}_{i=1}^N$
\For{$j=1, \dots, T$}
    \State Compute rewards $\mathbf{w} = [\hat{r}(x_{1:t}^{(1)}), \dots, \hat{r}(x_{1:t}^{(N)})]$
    \State Compute softmax distribution $\theta = \mathrm{softmax}(\mathbf{w})$
    \State Sample reference particle $x_{1:t}^{\text{ref}} := x_{1:t}^{(j)}$ where $j \sim \mathbb{P}(j=i) = \theta_i$
    \State Initialize $N-1$ particles $\{x_1^{(i)} \sim p_M(\cdot \mid c)\}_{i=1}^{N-1}$
    \State $t \gets 1$
    \While{not all particles stop} 
        \State Update $\mathbf{w} = [\hat{r}(x_{1:t}^{(1)}), \dots, \hat{r}(x_{1:t}^{(N-1)}), \hat{r}(x_{1:t}^{\text{ref}})]$
        \State Compute softmax distribution $\theta = \mathrm{softmax}(\mathbf{w})$
        \State Sample indices $\{j_t^{(i)}\}_{i=1}^N  \sim \mathbb{P}_t(j=i) = \theta_i$
        \State Update the set of particles as $\{x_{1:t}^{(j_t^{(i)})}\}_{i=1}^N$
        \State Transition $\{x_{t+1}^{(i)} \sim p_M(\cdot \mid c, x_{t+1}^{(i)})\}_{i=1}^N$
        \State $t \gets t + 1$
    \EndWhile
\EndFor
\State \textbf{Return}: the set of particles in the end
\end{algorithmic}
\end{algorithm}

For a set of parallel chains with temperatures $T_1 > T_2 > \dots$, at each iteration, we swap the states of every pair of neighboring chains $k, k+1$ with the following probability

\begin{equation}\label{eq:pt-swap-prob}
A = \min\left(1, \frac{\pi_k(x^{(k+1)}) \pi_{k+1}(x^{(k)})}{\pi_k(x^{(k)}) \pi_{k+1}(x^{(k+1)})}\right),
\end{equation}

where $\pi_k, \pi_{k+1}$ are the two targets (with different temperatures) and $x_k, x_{k+1}$ are their states before swapping.

\begin{algorithm}[h]
\caption{Particle Gibbs with Parallel Tempering for Inference-Time Scaling}\label{alg:pt}
\begin{algorithmic}
\State \textbf{Input}: same as \algref{alg:pg} with the number of parallel chains $M$ and a list of temperature $T_1, \dots, T_M$
\For{$j=1, \dots, T$}
    \For{$k=1, \dots, M$}
        \If{$j = 1$}
            \State Run \algref{alg:pf} to get a set of particles $\{x_{1:t}^{(i)}\}_{i=1}^N$ for chain $k$
        \Else
            \State Initialize $N-1$ particles $\{x_1^{(i)} \sim p_M(\cdot \mid c)\}_{i=1}^{N-1}$
            \State $t \gets 1$
            \While{not all particles stop} 
                \State Update $\mathbf{w} = [\hat{r}(x_{1:t}^{(1)}), \dots, \hat{r}(x_{1:t}^{(N-1)}), \hat{r}(x_{1:t}^{\text{ref}})]$
                \State Compute softmax distribution $\theta = \mathrm{softmax}(\mathbf{w} / T_k)$
                \State Sample indices $\{j_t^{(i)}\}_{i=1}^N  \sim \mathbb{P}_t(j=i) = \theta_i$
                \State Update the set of particles as $\{x_{1:t}^{(j_t^{(i)})}\}_{i=1}^N$
                \State Transition $\{x_{t+1}^{(i)} \sim p_M(\cdot \mid c, x_{t+1}^{(i)})\}_{i=1}^N$
                \State $t \gets t + 1$
            \EndWhile
        \EndIf
        \State Compute rewards $\mathbf{w} = [\hat{r}(x_{1:t}^{(1)}), \dots, \hat{r}(x_{1:t}^{(N)})]$
        \State Compute softmax distribution $\theta = \mathrm{softmax}(\mathbf{w} / T_k)$
        \State Sample reference particle $x_{1:t}^{\text{ref}} := x_{1:t}^{(j)}$ where $j \sim \mathbb{P}(j=i) = \theta_i$
    \EndFor
    \For{$k=1, \dots, M-1$}
        \State Exchange the reference particle between chain $k$ and $k+1$ with probability according to \eqref{eq:pt-swap-prob}
    \EndFor
\EndFor
\State \textbf{Return}: $M$ set of particles in the end
\end{algorithmic}
\end{algorithm}

\newpage

\section{Particle Filtering Results with More Generator Models}\label{app-moreResultsMoreModels}

 \vspace{-3.5em}
Below, we show further results using particle filtering to inference scale a wider variety of generator models. 

\begin{table}[h!]
\vspace{-6em}
\centering
\begin{tabular}{lr|cc}
\hline
\multicolumn{1}{c}{\textbf{Model}} & \multicolumn{1}{c|}{\textbf{Method}}                      & \textbf{MATH500}                     & \textbf{AIME2024}                    \\ \hline
Llama-3.2-1B-Instruct              & Greedy                                                    & 26.8                                 & 0.0                                  \\
\rowcolor[HTML]{C1DEB8} 
                                   & \textbf{Particle Filtering (Ours)}                        & \textbf{59.6}                        & \textbf{10.0}                        \\ \hline
Llama-3.2-8B-Instruct              & Greedy                                                    & 49.9                                 & 6.6                                  \\
\rowcolor[HTML]{C1DEB8} 
                                   & \textbf{Particle Filtering (Ours)}                        & \textbf{74.4}                        & \textbf{16.6}                        \\ \hline
phi-4                              & Greedy                                                    & 79.8                                 & 16.6                                 \\
\rowcolor[HTML]{C1DEB8} 
{\color[HTML]{000000} }            & {\color[HTML]{000000} \textbf{Particle Filtering (Ours)}} & {\color[HTML]{000000} \textbf{83.6}} & {\color[HTML]{000000} \textbf{26.6}} \\ \hline
Mistral-Small-24B-Instruct-2501    & Greedy                                                    & 69.2                                 & 10                                   \\
\rowcolor[HTML]{C1DEB8} 
                                   & \textbf{Particle Filtering (Ours)}                        & \textbf{83.4}                        & \textbf{23.3}                        \\ \hline
\end{tabular}
\vspace{.5em}
\caption{Performance of LLMs on MATH500 and AIME 2024 using greedy decoding and Particle Filtering (ours). Particle Filtering is run with 64 generations per problem.}
\label{tab:model_performance_neurips}
\end{table}

\section{Inference Prompt Template}\label{app:example}

\begin{tcolorbox}[colback=gray!10, colframe=gray!80, sharp corners, title=Evaluation System Prompt]\label{prompt_template}
\begin{verbatim}
Solve the following math problem efficiently and clearly:
    - For simple problems (2 steps or fewer):  
    Provide a concise solution with minimal explanation.

    - For complex problems (3 steps or more):
    Use this step-by-step format:
    
    ## Step 1: [Concise description]
    [Brief explanation and calculations]

    ## Step 2: [Concise description]
    [Brief explanation and calculations]

Regardless of the approach, always conclude with:

Therefore, the final answer is: $\boxed{answer}$. I hope it is correct.

Where [answer] is just the final number or expression that solves the problem.
\end{verbatim}
\end{tcolorbox}

\begin{tcolorbox}[colback=gray!10, colframe=gray!80, sharp corners, title=PRM Input Format]\label{code:prm}
\begin{verbatim}
## Step 1: [Concise description]
[Brief explanation and calculations]
<reward_token>
## Step 2: [Concise description]
[Brief explanation and calculations]
<reward_token>
## Step 3: [Concise description]
[Brief explanation and calculations]
<reward_token>
\end{verbatim}
\end{tcolorbox}

\begin{tcolorbox}[colback=gray!10, colframe=gray!80, sharp corners, title=ORM Input Format]\label{code:orm}
\begin{verbatim}
## Step 1: [Concise description]
[Brief explanation and calculations]
## Step 2: [Concise description]
[Brief explanation and calculations]
## Step 3: [Concise description]
[Brief explanation and calculations]
<reward_token>
\end{verbatim}
\end{tcolorbox}

\section{Evaluation details}\label{app:eval}

\paragraph{Parsing and scoring}
Following prior work on mathematical reasoning benchmarks \citep{qwen_2_5}, we apply their heuristic-based parsing and cleaning techniques to robustly extract the boxed expression. These heuristics handle spacing variations, formatting inconsistencies, and other artifacts in model outputs. For answer verification, we follow \citep{beeching2024scalingtesttimecompute}, converting responses to canonical form. Ground truth and generated answers are transformed from LaTeX into SymPy expressions, simplified for normalization, and converted back to LaTeX.
Exact match is determined using two criteria: numerical equality, where expressions evaluate to the same float, and symbolic equality, where they are algebraically equivalent in SymPy \citep{beeching2024scalingtesttimecompute}. Accuracy is computed as the fraction of problems where the generated answer exactly matches the ground truth.


\newpage

\section*{NeurIPS Paper Checklist}


\begin{enumerate}

\item {\bf Claims}
    \item[] Question: Do the main claims made in the abstract and introduction accurately reflect the paper's contributions and scope?
    \item[] Answer: \answerYes{} 
    \item[] Justification: We show experiments that directly address and back up every claim we make in the abstract and introduction. Please see the Evaluation section for our empirical results. 

\item {\bf Limitations}
    \item[] Question: Does the paper discuss the limitations of the work performed by the authors?
    \item[] Answer: \answerYes{} 
    \item[] Justification: We discuss several limitations of our work in the conclusion section. For clarity, we copy them here: "However, inference-time scaling comes with computational challenges. Hosting and running a reward model often introduces high latency, making the process more resource-intensive. For smaller models, prompt engineering is often required to ensure outputs adhere to the desired format. Finally, hyperparameters like budget are problem-dependent and may require tuning across domains."


\item {\bf Theory assumptions and proofs}
    \item[] Question: For each theoretical result, does the paper provide the full set of assumptions and a complete (and correct) proof?
    \item[] Answer: \answerYes{} 
    \item[] Justification: We have 1 theorem in our paper, and its proof is provided in Appendix \ref{app-proof-unbiasedness}. 

    \item {\bf Experimental result reproducibility}
    \item[] Question: Does the paper fully disclose all the information needed to reproduce the main experimental results of the paper to the extent that it affects the main claims and/or conclusions of the paper (regardless of whether the code and data are provided or not)?
    \item[] Answer: \answerYes{} 
    \item[] Justification: We include full details of how exactly our overall methodology work. We also provide key details into the hyperparameter selection and ablation process, which significantly helps reproducability. We also provide details into several reward model aggregation techniques, as well as why we chose the one we chose, in our paper. We also include details on several different Process Reward Models and include ablations showing why we chose the one we chose. Our experiments are reproducible. 

\item {\bf Open access to data and code}
    \item[] Question: Does the paper provide open access to the data and code, with sufficient instructions to faithfully reproduce the main experimental results, as described in supplemental material?
    \item[] Answer: \answerNo{} 
    \item[] Justification: We do not currently include code as to not break author confidentiality. However, we will completely open source our code upon acceptance of the paper to encourage as many people as possible to use our work. 

\item {\bf Experimental setting/details}
    \item[] Question: Does the paper specify all the training and test details (e.g., data splits, hyperparameters, how they were chosen, type of optimizer, etc.) necessary to understand the results?
    \item[] Answer: \answerYes{}
    \item[] Justification: We provide full details on all datasets we used, as well as their citations. We list exactly which models and versions we used as generator models. We also provide key details into the hyperparameter selection and ablation process, which significantly helps reproducability. We also provide details into several reward model aggregation techniques, as well as why we chose the one we chose, in our paper. We also include details on several different Process Reward Models and include ablations showing why we chose the one we chose. Our experiments are reproducible. 

\item {\bf Experiment statistical significance}
    \item[] Question: Does the paper report error bars suitably and correctly defined or other appropriate information about the statistical significance of the experiments?
    \item[] Answer: \answerNo{} 
    \item[] Justification: Due to computational limitations on an academic budget, we were not able to run every single experiment multiple times to produce accurate and fair error bars across every experiment in the paper. However, we have run many experiments several times during the research and development process (both several times and by several different people) and are very confident in our results. 

\item {\bf Experiments compute resources}
    \item[] Question: For each experiment, does the paper provide sufficient information on the computer resources (type of compute workers, memory, time of execution) needed to reproduce the experiments?
    \item[] Answer: \answerNo{} 
    \item[] Justification: Although we include information about the computational burden of running inference time scaling experiments, we do not provide formal information about which exact computational resources we used, as we used different numbers of GPUs for different experiments. That being said, we have developed a (to be released upon acceptance) open source library for our work that is able to be used completely off the shelf and contains very simple information about how to run it. 
    
\item {\bf Code of ethics}
    \item[] Question: Does the research conducted in the paper conform, in every respect, with the NeurIPS Code of Ethics \url{https://neurips.cc/public/EthicsGuidelines}?
    \item[] Answer: \answerYes{} 
    \item[] Justification: We can confirm that, in every respect, we do not violate the NeurIPS Code of Ethics. Our research does not have negative societal consequences, nor does it involve human subjects. We do not use any private or sensitive data. 

\item {\bf Broader impacts}
    \item[] Question: Does the paper discuss both potential positive societal impacts and negative societal impacts of the work performed?
    \item[] Answer: \answerYes{} 
    \item[] Justification: Our work does not have any negative societal consequences. We discuss the positive impacts of inference scaling, as it opens up higher level language model performance to those who are only able to access smaller models. 
    
\item {\bf Safeguards}
    \item[] Question: Does the paper describe safeguards that have been put in place for responsible release of data or models that have a high risk for misuse (e.g., pretrained language models, image generators, or scraped datasets)?
    \item[] Answer: \answerNo{} 
    \item[] Justification: We do not release any data or models of our own. Instead, we only use off-the-shelf open source models, and therefore there are no possibilities of misuse. 

\item {\bf Licenses for existing assets}
    \item[] Question: Are the creators or original owners of assets (e.g., code, data, models), used in the paper, properly credited and are the license and terms of use explicitly mentioned and properly respected?
    \item[] Answer: \answerYes{} 
    \item[] Justification: Our work only uses open source models, and we cite every model that we use. Therefore, all creators of the original models are credited in this work. 

\item {\bf New assets}
    \item[] Question: Are new assets introduced in the paper well documented and is the documentation provided alongside the assets?
    \item[] Answer:  \answerNA{}.
    \item[] Justification: We do not release any new assets in this paper - instead, we discuss how to enhance the performance of already existing open-sourced models. 

\item {\bf Crowdsourcing and research with human subjects}
    \item[] Question: For crowdsourcing experiments and research with human subjects, does the paper include the full text of instructions given to participants and screenshots, if applicable, as well as details about compensation (if any)? 
    \item[] Answer: \answerNA{}
    \item[] Justification: Our research does not include any human subject experiments or crowdsourcing experiments. 

\item {\bf Institutional review board (IRB) approvals or equivalent for research with human subjects}
    \item[] Question: Does the paper describe potential risks incurred by study participants, whether such risks were disclosed to the subjects, and whether Institutional Review Board (IRB) approvals (or an equivalent approval/review based on the requirements of your country or institution) were obtained?
    \item[] Answer: \answerNA{}.
    \item[] Justification: Our work does not include any human subjects, and we did not need IRB approvals. 

\item {\bf Declaration of LLM usage}
    \item[] Question: Does the paper describe the usage of LLMs if it is an important, original, or non-standard component of the core methods in this research? Note that if the LLM is used only for writing, editing, or formatting purposes and does not impact the core methodology, scientific rigorousness, or originality of the research, declaration is not required.
    \item[] Answer: \answerNA{}
    \item[] Justification: LLM usage did not impact the core methodology, scientific rigorousness, or originality of the research. 

\end{enumerate}

\end{document}